\documentclass[runningheads]{llncs}

\usepackage{eccv}

\usepackage{multirow}
\usepackage{colortbl}

\usepackage{eccvabbrv}

\usepackage{graphicx}
\usepackage{booktabs}

\usepackage[accsupp]{axessibility}  
\usepackage{amsmath}
\DeclareMathOperator*{\argmax}{arg\,max}

\definecolor{Gray}{gray}{0.85}
\newcolumntype{a}{>{\columncolor{Gray}}c}

\usepackage{hyperref}

\usepackage{orcidlink}

\begin{document}

\title{SA-DVAE: Improving Zero-Shot Skeleton-Based Action Recognition by Disentangled Variational Autoencoders}

\titlerunning{Semantic Alignment via Disentangled Variational Autoencoders}

\author{
    Sheng-Wei Li\inst{1}\orcidlink{0009-0002-9091-8036}\and
    Zi-Xiang Wei\inst{2}\orcidlink{0009-0002-8214-3226}\and
    Wei-Jie Chen\inst{2}\orcidlink{0009-0001-0557-8106}\and
    Yi-Hsin Yu\inst{2}\orcidlink{0009-0008-6707-8589}\and
    Chih-Yuan Yang\inst{3,4}\orcidlink{0000-0002-8989-501X}\and
    Jane Yung-jen Hsu\inst{2,3}\orcidlink{0000-0002-2408-4603}
}

\authorrunning{S.-W.~Li et al.}

\institute{
Graduate Institute of Networking and Multimedia, National Taiwan University
\and
Department of Computer Science and Information Engineering, National Taiwan University 
\and
Department of Artificial Intelligence, Chang Gung University 
\and
Artificial Intelligence Research Center, Chang Gung University 
\\
\email{\{r11944004,r12922147,r12922051,r12922220,yjhsu\}@csie.ntu.edu.tw, cyyang@cgu.edu.tw}
}
\maketitle

\begin{abstract}
    Existing zero-shot skeleton-based action recognition methods utilize projection networks to learn a shared latent space of skeleton features and semantic embeddings. The inherent imbalance in action recognition datasets, characterized by variable skeleton sequences yet constant class labels, presents significant challenges for alignment. To address the imbalance, we propose SA-DVAE---Semantic Alignment via Disentangled Variational Autoencoders, a method that first adopts feature disentanglement to separate skeleton features into two independent parts---one is semantic-related and another is irrelevant---to better align skeleton and semantic features. We implement this idea via a pair of modality-specific variational autoencoders coupled with a total correction penalty. We conduct experiments on three benchmark datasets: NTU RGB+D, NTU RGB+D 120 and PKU-MMD, and our experimental results show that SA-DAVE produces improved performance over existing methods. The code is available at \url{https://github.com/pha123661/SA-DVAE}.
\end{abstract}

\section{Introduction}
\label{sec:introduction}

Action recognition is a long-standing active research area because it is challenging and has a wide range of applications like surveillance, monitoring, and human-computer interfaces. Based on input data types, there are several lines of studies on human action recognition: image-based, video-based, depth-based, and skeleton-based. In this paper, we focus on the skeleton-based action recognition, which is enabled by the advance in pose estimation~\cite{sun2019hrnet, yuan2021hrformer} and sensor~\cite{zhang2012microsoft, keselman2017realsense} technologies, and has emerged as a viable alternative to video-based action recognition due to its resilience to variations in appearance and background. Some existing skeleton-based action recognition methods already achieve remarkable performance on large-scale action recognition datasets~\cite{shahroudy2016ntu60, liu2019ntu120, liu2017pkummd} through supervised learning, but labeling data is expensive and time-consuming. For the cases where training data are difficult to obtain or prevented by privacy issues, zero-shot learning (ZSL) offers an alternative solution by recognizing unseen actions through supporting information such as the names, attributes, or descriptions of the unseen classes. Therefore, zero-shot learning has multiple types of input data and aims to learn an effective way of dealing with those data representations. For skeleton-based zero-shot action recognition, several methods have been proposed to align skeleton features and text features in the same space.

\begin{figure}[t]
    \centering
    \noindent\resizebox{\textwidth}{!}{
    \includegraphics[]{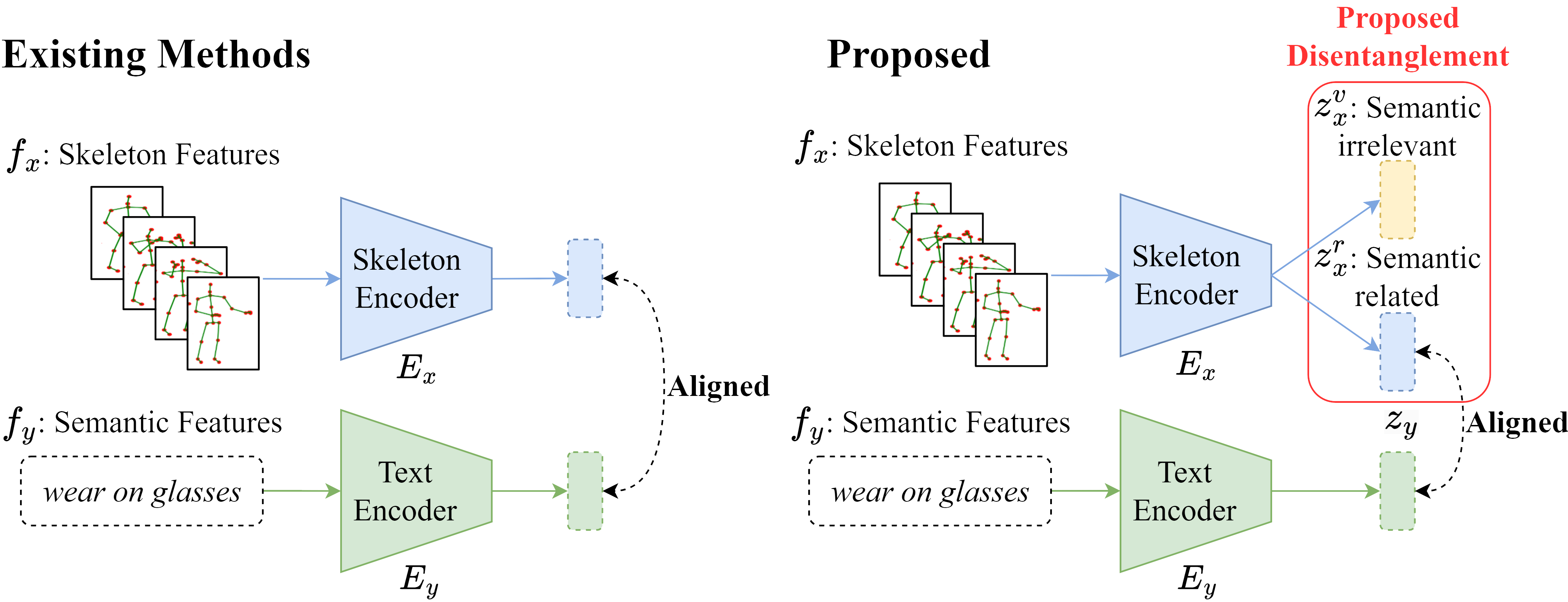}}
    \caption{Comparison with existing methods. Our method is the first to apply feature disentanglement to the problem of skeleton-based zero-shot action recognition. All existing methods directly align skeleton features with textual ones, but ours only aligns a part of semantic-related skeleton features with the textual ones.}
    \label{fig:highlight-comparison}
\end{figure}

However, to the best of our knowledge, all existing methods assume that the group of skeleton sequences are well captured and highly consistent so their ideas mainly focus on how to semantically optimize text representation. After carefully examining the source videos in two widely used benchmark datasets NTU RGB+D and PKU-MMD, we found the assumption is questionable. We observe that for some labels, the camera positions and actors' action differences do bring in significant noise. To address this observation, we seek an effective way to deal with the problem. Inspired by an existing ZSL method~\cite{chen2021tcpenalty} which shows semantic-irrelevant features can be separated from semantic-related ones,
we propose SA-DVAE for skeleton-based action recognition. SA-DVAE tackles the generalization problem by disentangling the skeleton latent feature space into two components: a semantic-related term and a semantic-irrelevant term as shown in \cref{fig:highlight-comparison}. This enables the model to learn more robust and generalizable visual embeddings by focusing solely on the semantic-related term for action recognition. In addition, SA-DVAE implements a learned total correlation penalty that encourages independence between the two factorized latent features and minimizes the shared information captured by the two representations. This penalty is realized by an adversarial discriminator that aims to estimate the lower bound of the total correlation between the factorized latent features.

The contributions of our paper are as follows:
\begin{itemize}
    \item We propose a novel SA-DVAE method. By disentangling the latent feature space into semantic-related and irrelevant terms, the model addresses the asymmetry existing in action recognition datasets and improves the generalization capability.
    \item We leverage an adversarial total correlation penalty to encourage independence between the two factorized latent features.
    \item We conduct extensive experiments that show SA-DVAE achieves state-of-the-art performance on the ZSL and generalized zero-shot learning (GZSL) benchmarks of the NTU RGB+D 60, NTU RGB+D 120, and PKU-MMD datasets.
\end{itemize}

\section{Related Work}
\label{sec:related}
The proposed SA-DAVE method covers two research fields: zero-shot learning and action recognition, and it uses feature disentanglement to deal with skeleton data noise. Here we discuss the most related research reports in the literature.

\subsubsection{Skeleton-Based Zero-Shot Action Recognition.}
ZSL aims to train a model under the condition that some classes are unseen during training. The more challenging GZSL expands the task to classify both seen and unseen classes during testing~\cite{pourpanah2022reviewgzsl}. ZSL relies on semantic information to bridge the gap between seen and unseen classes.

Existing methods address the skeleton and text zero-shot action recognition problem by constructing a shared space for both modalities. ReViSE~\cite{hubert2017revise} learns autoencoders for each modality and aligns them by minimizing the maximum mean discrepancy loss between the latent spaces. Building on the concept of feature generation, CADA-VAE~\cite{schonfeld2019cadavae} employs variational autoencoders (VAEs) for each modality, aligning the latent spaces through cross-modal reconstruction and minimizing the Wasserstein distance between the inference models. These methods then learn classifiers on the shared space to conduct classification.

SynSE~\cite{gupta2021synse} and JPoSE~\cite{wray2019JPoSE} are two methods that leverage part-of-speech (PoS) information to improve the alignment between text descriptions and their corresponding visual representations. SynSE extends CADA-VAE by decomposing text descriptions by PoS tags, creating individual VAEs for each PoS label, and aligning them in the skeleton space. Similarly, JPoSE~\cite{wray2019JPoSE} learns multiple shared latent spaces for each PoS label using projection networks. JPoSE employs uni-modal triplet loss to maintain the neighborhood structure of each modality within the shared space and cross-modal triplet loss to align the two modalities.

On the other hand, SMIE~\cite{zhou2023smie} focuses on maximizing mutual information between skeleton and text feature spaces, utilizing a Jensen-Shannon Divergence estimator trained with contrastive learning. It also considers temporal information in action sequences by promoting an increase in mutual information as more frames are observed.

While JPoSE and SynSE demonstrate the benefits of incorporating PoS information, they rely heavily on it and require additional PoS tagging effort. Furthermore, the two methods neglect the inherent asymmetry between modalities, aligning semantic-related and irrelevant terms to the semantic features and missing the chance to improve recognition accuracy further. In contrast, our approach uses simple class labels without the need of PoS tags, and uses only semantic-related skeleton information to align text data.

\noindent\textbf{Feature Disentanglement in Generalized Zero-Shot Learning.}
Feature disentanglement refers to the process of separating the underlying factors of variation in data~\cite{bengio2013featuredisentanglement}. 
Because methods of zero-shot learning are sensitive to the quality of both visual and semantic features, feature disentanglement serves as an effective approach to scrutinize either visual or semantic features, as well as addressing the domain shift problem~\cite{pourpanah2022reviewgzsl}, thereby generating more robust and generalized representations.

SDGZSL~\cite{chen2021tcpenalty} decomposes visual embeddings into semantic-consistent and semantic-unrelated components using shared class-level attributes, and learns an additional relation network to maximize compatibility between semantic-consistent representations and their corresponding semantic embeddings. This approach is motivated by the transfer of knowledge from intermediate semantics (e.g., class attributes) to unseen classes. In contrast, SA-DVAE addresses the inherent asymmetry between the text and skeleton modalities, enabling the direct use of text descriptions instead of relying on predefined class attributes.

\begin{figure}[t]
    \centering
    \noindent\resizebox{\textwidth}{!}{
    \includegraphics[]{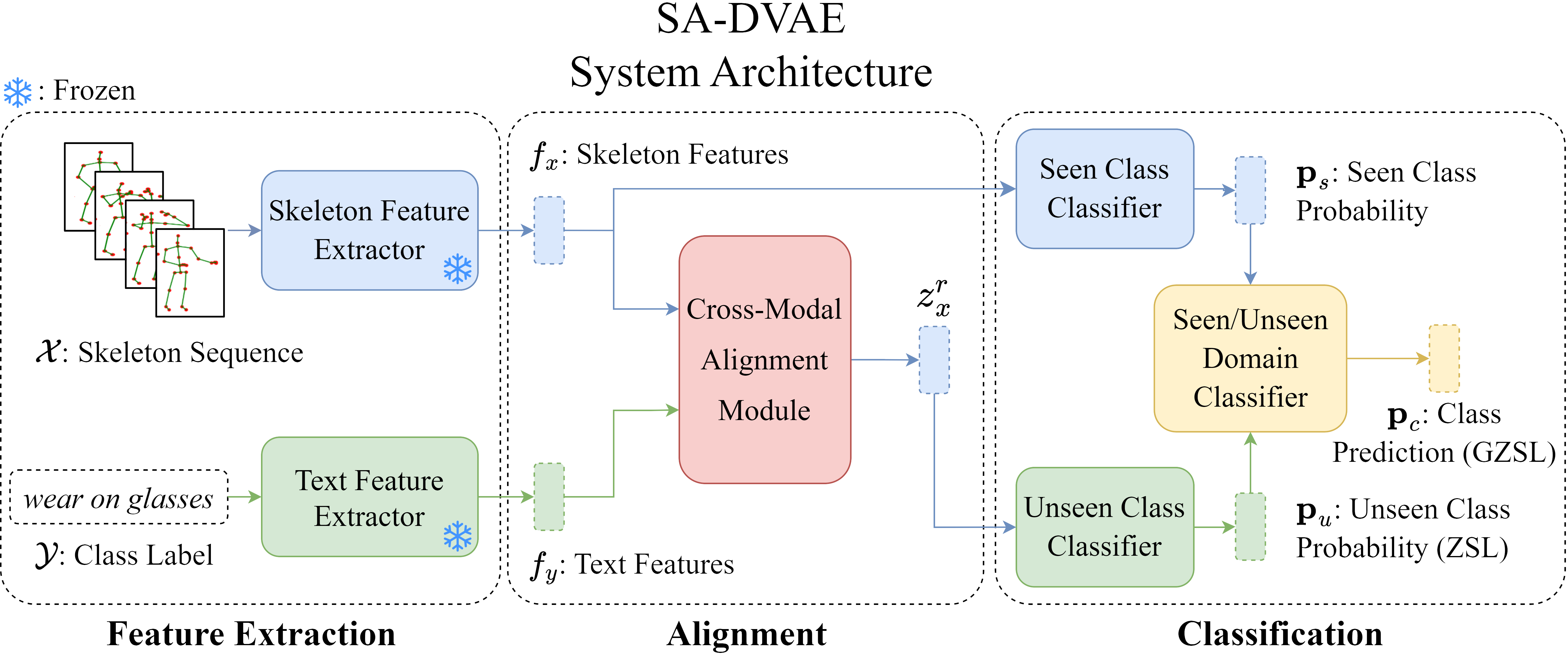}
    }
    \caption{System Architecture of SA-DVAE. Initially, the feature extractors are employed to extract features. Subsequently, the cross-modal alignment module aligns the two modalities and generates semantic-related unseen skeleton features ($z^r_x$). These generated features are utilized to train classifiers.}
    \label{fig:system-architecture}
\end{figure}

\section{Methodology}
\label{sec:method}

We show the overall architecture of our method as \cref{fig:system-architecture}, which consists of three main components: a) two modality-specific feature extractors, b) a cross-modal alignment module, and c) three classifiers for seen/unseen actions and their domains. The cross-modal alignment module learns a shared latent space via cross-modality reconstruction, where feature disentanglement is applied to prioritize the alignment of semantic-related information ($z^r_x$ and $z_y$). To improve the effectiveness of the disentanglement, we use a discriminator as an adversarial total correlation penalty between the disentangled features.

\noindent\textbf{Problem Definition.}
Let $\mathcal{D}$ be a skeleton-based action dataset consisting of a skeleton sequences set $\mathcal{X}$ and a label set $\mathcal{Y}$, in which a label is a piece of text description.
The $\mathcal{X}$ is split into a seen and unseen subset $\mathcal{X}_s$ and $\mathcal{X}_u$ where we can only use $\mathcal{X}_s$ and $\mathcal{Y}$ to train a model to classify $x \in \mathcal{X}_u$.
By definition, there are two types of evaluation protocols. The GZSL one asks to predict the class of $x$ among all classes $\mathcal{Y}$, and the ZSL only among $\mathcal{Y}_u = \{y_i : x_i \in \mathcal{X}_u\}$.

\begin{figure}[t]
    \centering
    \noindent\resizebox{\textwidth}{!}{
    \includegraphics[]{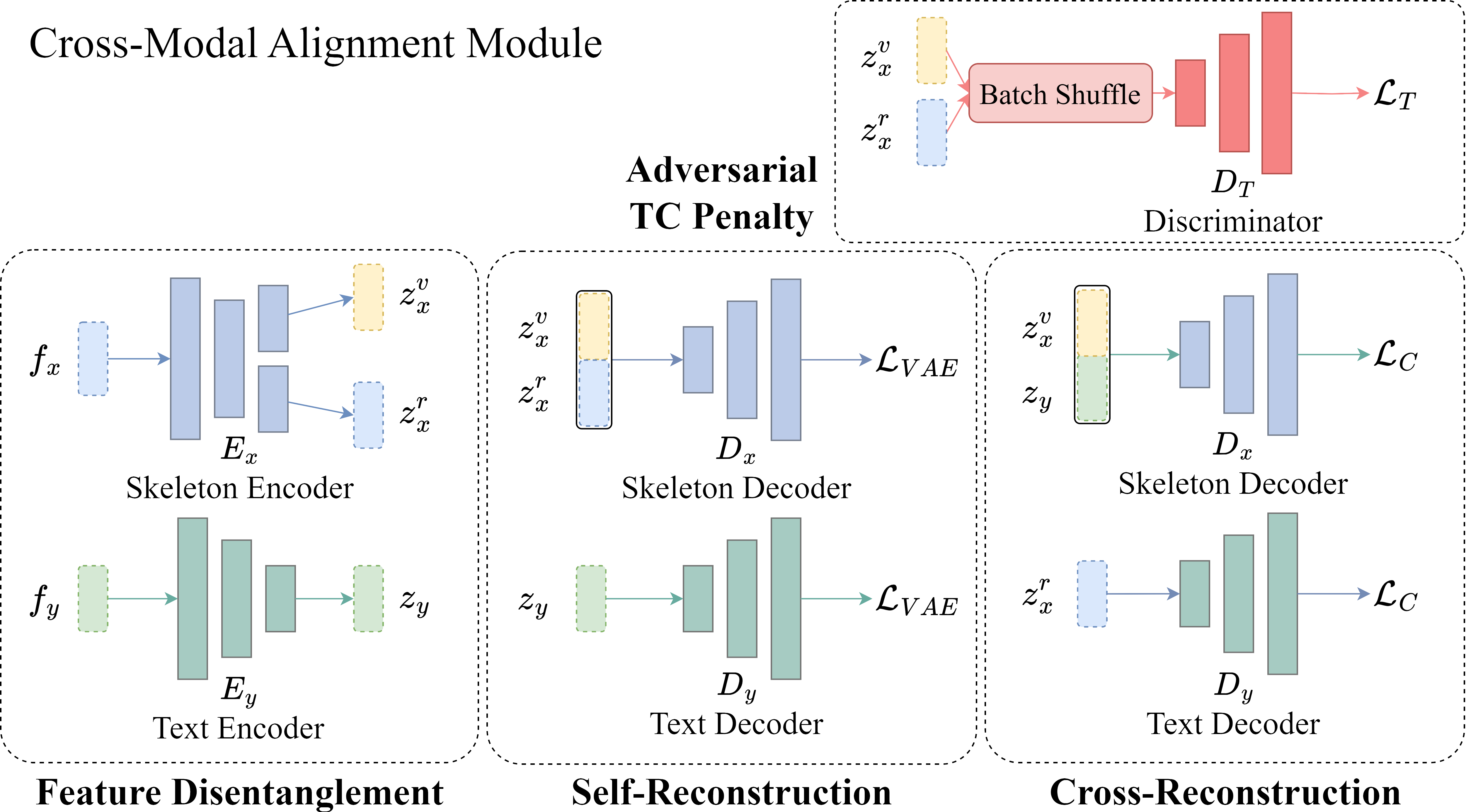}}
    \caption{Cross-Modal Alignment Module. This module serves two primary tasks: latent space construction through self-reconstruction and cross-modal alignment via cross-reconstruction. The skeleton features are disentangled into semantic-related ($z^r_x$) and irrelevant ($z^v_x$) factors.}
    \label{fig:cross-modal-alignment-module}
\end{figure}

\noindent\textbf{Cross-Modal Alignment Module.}
\label{sec:cross-modal-alignment-module}
We train a skeleton representation model (Shift-GCN~\cite{cheng2020shiftgcn} or ST-GCN~\cite{yan2018stgcn}, depending on experimental settings) on the seen classes using standard cross-entropy loss. This model extracts our skeleton features, denoted as $f_x$. We use a pre-trained language model (Sentence-BERT~\cite{reimers2019sentence} or CLIP~\cite{radford2021clip}) to extract our label's text features, denoted as $f_y$.
Because $f_x$ and $f_y$ belong to two unrelated modalities, we train two modality-specific VAEs to adjust $f_x$ and $f_y$ for our recognition task and illustrate their data flow in \cref{fig:cross-modal-alignment-module}. Our encoders $E_x$  and $E_y$ transform $f_x$ and $f_y$ into representations $z_x$ and $z_y$ in a shared latent space via the reparameterization trick~\cite{kingma2013reparameterization}. To optimize the VAEs, we introduce a loss as the form of the Evidence Lower Bound
\begin{align}
    \mathcal{L} = \mathbb{E}_{q_\phi(z|f)} [\log p_\theta(f|z)] - \beta D_{\it KL}(q_\phi(z|f) \| p_\theta(z)),
\end{align}
where $\beta$ is a hyperparameter, $f$ and $z$ are the observed data and latent variables, the first term is the reconstruction error, and the second term is the Kullback-Leibler divergence between the approximate posterior $q(z|f)$ and $p(z)$. The hyperparameter $\beta$ balances the quality of reconstruction with the alignment of the latent variables to a prior distribution~\cite{higgins2016betavae}. We use multivariate Gaussian as the prior distribution.

\noindent\textbf{Feature Disentanglement.} 
We observe that although two skeleton sequences belong to the same class (\ie they share the same text description), their movement varies substantially due to stylistic factors such as actors' body shapes and movement ranges, and cameras' positions and view angles. To the best of our knowledge, existing methods never address this issue. For example, Zhou \etal.~\cite{zhou2023smie} and Gupta \etal~\cite{gupta2021synse} neglect this issue and force $f_x$ and $f_y$ to be aligned. Therefore, we propose to tackle the problem of inherent asymmetry between the two modalities to improve the recognition performance.

We design our skeleton encoder $E_x$ as a two-head network, of which one head generates a semantic-related latent vector $z^r_x$ and the other generates a semantic-irrelevant vector $z^v_x$. We assume each of $z^r_x$ and $z^v_x$ has its own multivariant normal distribution $N(\mu^r_x, \Sigma^r_x)$ and $N(\mu^v_x, \Sigma^v_x)$, and our text encoder $E_y$ generates a latent feature $z_y$, which also has a multivariant normal distribution $N(\mu_y, \Sigma_y)$.

Let $z_x = z^v_x \oplus z^r_x$ where $\oplus$ means concatenation.
We define the losses for the VAEs as
\begin{equation}
\begin{aligned}
    \mathcal{L}_x & = \mathbb{E}_{q_\phi(z_x|f_x)} [\log p_\theta(f_x|z_x)]                     \\
                  & - \beta_x D_{\it KL}(q_\phi(z^r_x|f_x) || p_\theta(z^r_x))       \\
                  & - \beta_x D_{\it KL}(q_\phi(z^v_x|f_x) || p_\theta(z^v_x)), 
\end{aligned}
\end{equation}
\begin{equation}
\mathcal{L}_y = \mathbb{E}_{q_\phi(z_y|f_y)} [\log p_\theta(f_y|z_y)]
                 - \beta_y D_{\it KL}(q_\phi(z_y|f_y) || p_\theta(z_y)),
\end{equation}
where $\beta_x$ and $\beta_y$ are hyperparameters, $p_\theta(z^r_x)$, $p_\theta(z^v_x)$, $p_\theta(f_x|z_x)$, $p_\theta(z_y)$, and $p_\theta(f_y|z_y)$ are the probabilities of their presumed distributions, $q_\phi(z_x|f_x)$, $q_\phi(z^r_x|f_x)$ and $q_\phi(z^v_x|f_x)$ are the probabilities calculated through our skeleton encoder $E_x$, and $q_\phi(z_y|f_y)$ is the one through our text encoder $E_y$.
We set the overall VAE loss as
\begin{equation}
    \mathcal{L}_{\it VAE} = \mathcal{L}_x + \mathcal{L}_y.
\end{equation}

\begin{figure}[h]
    \centering
    \begin{subfigure}[b]{0.4585\linewidth}
        \includegraphics[width=\linewidth]{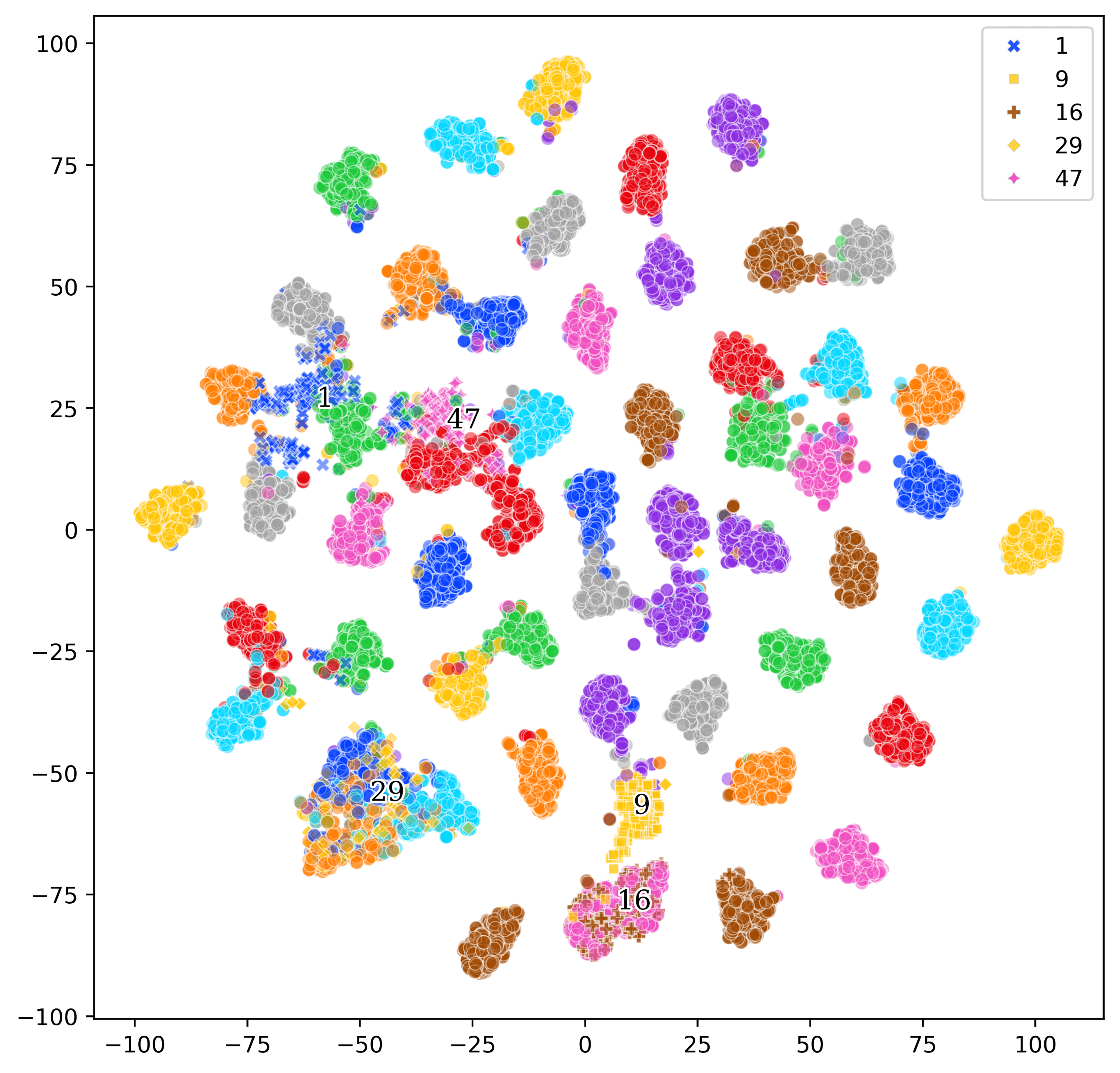}
        \caption{t-SNE visualization of $z^r_x$.}
        \label{subfig:murx}
    \end{subfigure}\hfill%
    \begin{subfigure}[b]{0.4585\linewidth}
        \includegraphics[width=\linewidth]{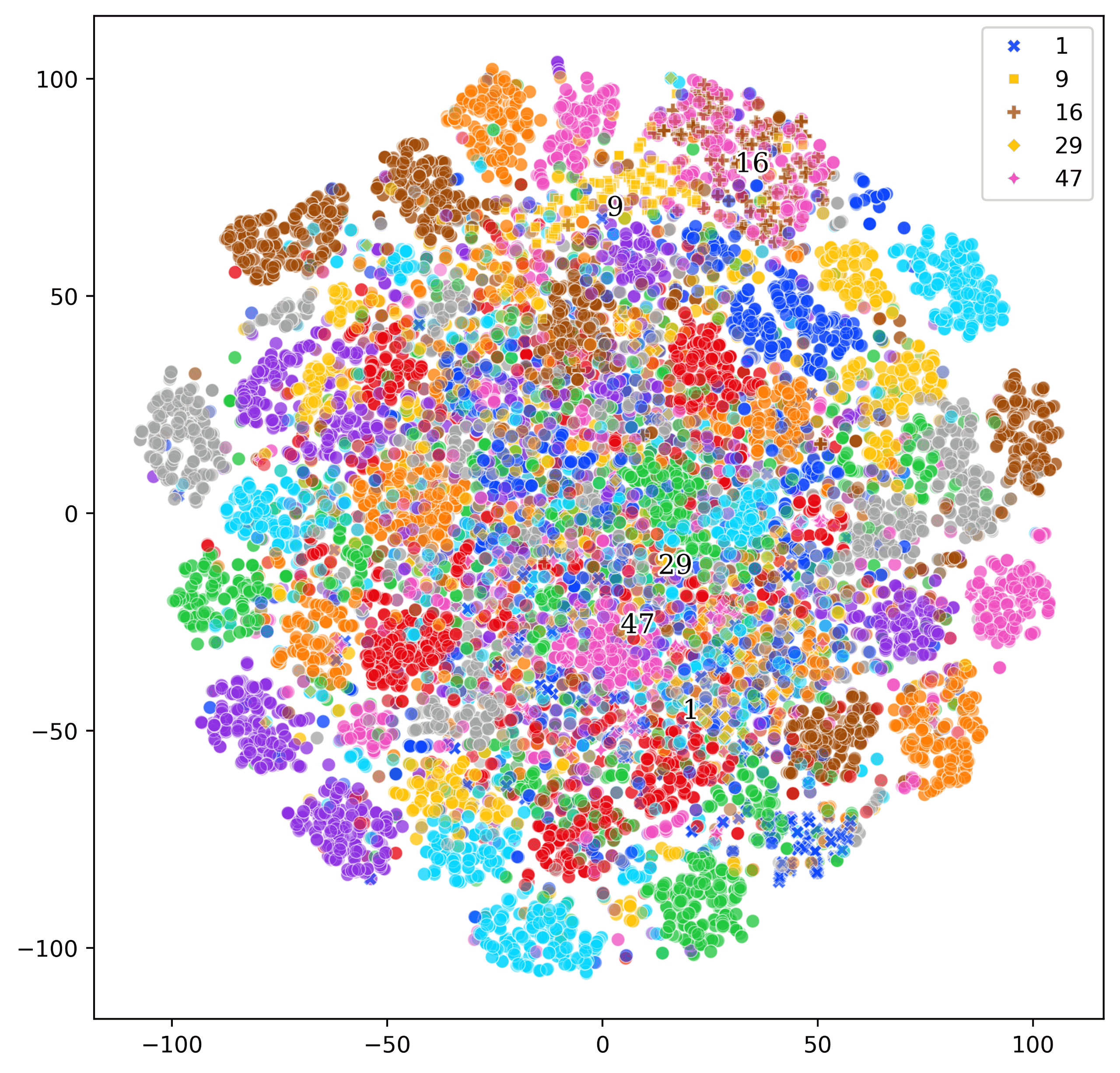}
        \caption{t-SNE visualization of $z^v_x$.}
        \label{subfig:muvx}
    \end{subfigure}%
    \caption{t-SNE visualizations of $z^r_x$ and $z^v_x$. Best viewed in color.}
    \label{fig:overall}
\end{figure}

To better understand our method, we present the t-SNE visualization of the semantic-related and semantic-irrelevant terms, $z^r_x$ and $z^v_x$ in \cref{fig:overall}. \Cref{subfig:murx} displays the t-SNE results for $z^r_x$, showing clear class clusters that demonstrate effective disentanglement. In contrast, \Cref{subfig:muvx} shows the t-SNE results for $z^v_x$, where class separation is less distinct. This indicates that while our method effectively clusters related semantic features, the irrelevant features remain more dispersed as they contain instance-specific information.

\noindent\textbf{Cross-Alignment Loss.} 
Because we want our latent text features $z_y$ to align with semantic-related skeleton features $z^r_x$ only, regardless of the semantic-irrelevant features $z^v_x$, we regulate them by setting up a cross-alignment loss 
\begin{equation}
    \label{eq:cross-alignment}
    \mathcal{L}_{C} = \lVert D_y(z^r_x) - f_y \rVert_2^2 + \lVert D_x(z^v_x \oplus z_y) - f_x \rVert_2^2
\end{equation}
to train our VAEs for skeleton and text respectively. This loss enforces skeleton features to be reconstructable from text features and vice versa. To reconstruct skeleton features from text features, $z^v_x$ is employed to incorporate necessary style information to mitigate the information gap between the class label and the skeleton sequence.

\noindent\textbf{Adversarial Total Correlation Penalty.}
We expect the features $z^r_x$ and $z^v_x$ to be statistically independent, so we impose an adversarial total correlation penalty~\cite{chen2021tcpenalty} on them. We train a discriminator $D_T$ to predict the probability of a given latent skeleton vector $z^v_x \oplus z^r_x$ whether the $z^v_x$ and $z^r_x$ come from the same skeleton feature $f_x$. In the ideal case, $D_T$ will return 1 if $z^v_x$ and $z^r_x$ are generated together, and 0 otherwise. To train $D_T$, we design a loss
\begin{equation}
    \mathcal{L}_T = \log D_T(z_x) + \log(1 - D_T(\Tilde{z}_x)),
\end{equation}
where $\Tilde{z}_x$ is an altered feature vector. We create $\tilde{z}_x$ as the following steps.
From a batch of $N$ training samples, our encoder $E_x$ generates $N$ pairs of $z_{x,i}^v$ and $z_{x,i}^r$, $i = 1 \dots N$. 
We randomly permute the indices $i$ of $z_{x,i}^v$ but keep $z_{x,i}^r$ unchanged, and then we concatenate them as $\Tilde{z}_x$. 
$D_T$ is trained to maximize $L_T$, while $E_x$ is adversarially trained to minimize it. 
This training process encourages the encoder to generate latent representations that are independent.
Combining the three losses, we set the overall loss 
\begin{equation}
    \mathcal{L} = \mathcal{L}_{\it VAE} + \lambda_1\mathcal{L}_C + \lambda_2\mathcal{L}_T,
\end{equation}
where we balance the three losses by hyperparameters $\lambda_1$ and $\lambda_2$.

\noindent\textbf{Seen, Unseen and Domain Classifier.}
Because there are two protocols, ZSL and GZSL, to evaluate a zero-shot recognition model, we use two different settings for the two protocols.
For the ZSL protocol, we only need to predict the probabilities of classes $\mathcal{Y}_u$ from a given skeleton sequence, so we propose
a classifier $C_u$ as a single-layer MLP (Multilayer Perception) with a softmax output layer yielding the probabilities to predict probabilities of classes $\mathcal{Y}_u$ from $z_y$ by
\begin{equation}
    \mathbf{p}_u = C_u(z_y) = C_u(E_y(f_y)),
\end{equation}
where dim($\mathbf{p}_u$) = $|\mathcal{Y}_u|$. 
During inference and given an unseen skeleton feature $f^u_x$, 
we get $z^u_x = E_x(f^u_x)$, separate $z^u_x$ into $z^{v,u}_x$ and $z^{r,u}_x$, 
and generate $\mathbf{p}_u = C_u(z^{r,u}_x)$ 
to predict its class as $y_{\hat{i}}$ and
\begin{equation}
\hat{i} = \argmax_{i = 1,\dots,|\mathcal{Y}_u|} p^i_u,
\end{equation}
where $p^i_u$ is the i-th probability value of $\mathbf{p}_u$.

For the GZSL protocol, we need to predict the probabilities of all classes in $\mathcal{Y} = \mathcal{Y}_u \cup \mathcal{Y}_s$ where $\mathcal{Y}_s = \{y_i : x_i \in \mathcal{X}_s\}$. We follow the same approach proposed by Gupta \etal~\cite{gupta2021synse} to use an additional class classifier $C_s$ for seen classes and a domain classifier $C_d$ to merge two arrays of probabilities.
Gupta \etal first apply Atzmon and Chechik's idea~\cite{atzmon2019adaptive} to a skeleton-based action recognition problem and outperform the typical single-classifier approach. The advantage of using dual classifiers is reported in a review paper~\cite{pourpanah2022reviewgzsl}.
Our $C_s$ is also a single-layer MLP with a softmax output layer like $C_u$, but it uses skeleton features $f_x$ rather than latent features to produce probabilities
\begin{equation}
    \mathbf{p}_s = C_s(f_x),
\end{equation}
where dim($\mathbf{p}_s$) = $|\mathcal{Y}_s|$.

We train $C_s$ and $C_u$ first, and then we freeze their parameters to train $C_d$, which is a logistic regression with an input vector $\mathbf{p}'_s \oplus \mathbf{p}_u$ where $\mathbf{p}'_s$ is the temperature-tuned~\cite{hinton2015distilling} top $k$-pooling result of $\mathbf{p}_s$ and the number $k$ = dim($\mathbf{p}_u$).
$C_d$ yields a probability value $p_d$ of whether the source skeleton belongs to a seen class.
We use the LBFGS algorithm~\cite{liu1989lbfgs_algo} to train $C_d$ and use it during inference to predict the probability of $x$ as
\begin{align}
    \mathbf{p}(y|x) = C_d(\mathbf{p}'_s \oplus \mathbf{p}_u) \mathbf{p}_s \oplus (1-C_d(\mathbf{p}'_s \oplus \mathbf{p}_u)) \mathbf{p}_u  = p_d \mathbf{p}_s \oplus (1-p_d) \mathbf{p}_u
\end{align}
and decide the class of $x$ as $y_{\hat{i}}$ and
\begin{equation}
\hat{i} = \argmax_{i = 1,\dots,|\mathcal{Y}|} p^i,
\end{equation}
where $p^i$ is the i-th probability value of $\mathbf{p}(y|x)$.

\section{Experiments}
\label{sec:experiments}

\noindent\textbf{Datasets.}
We conduct experiments on three datasets and show their statistics in Table~\ref{tab:dataset_statistics}. We adopt the cross-subject split, where half of the subjects are used for training and the other half for validation. We use NTU-60 and NTU-120 as synonyms for the NTU RGB+D and NTU RGB+D 120 datasets. Due to discrepancies in class labels between the official website\footnote{Official website: \url{https://rose1.ntu.edu.sg/dataset/actionRecognition/}} and the GitHub codebase\footnote{GitHub link: \url{https://github.com/shahroudy/NTURGB-D}} of NTU-60 and NTU-120 datasets (\eg the label of class 18 is ``put on glasses'' in their website but ``wear on glasses'' in GitHub), we follow existing methods by using the class labels provided in their codebase.

\begin{table}[t]
\centering
\caption{Statistics of datasets used in our experiments}
\begin{tabular}{lccccc}
\toprule
Name                                    & Class & Subject   & Joint  & Sample  & Camera View\\
\midrule
NTU RGB+D~\cite{shahroudy2016ntu60}     & 60     & 40       & 25     & 56,880  & 3\\
NTU RGB+D 120~\cite{liu2019ntu120}      & 120    & 106      & 25     & 114,480 & 3\\
PKU-MMD~\cite{liu2017pkummd}            & 51     & 66       & 25     & 28,443  & 3 \\
\bottomrule
\end{tabular}
\label{tab:dataset_statistics}
\end{table}

\noindent\textbf{Implementation Details.}
We implement the discriminator $D_T$ as a two-layer MLP with ReLU activation and a Sigmoid output layer, and the encoders $E_x$, $E_y$, decoders $D_x$, $D_y$, seen and unseen classifiers $C_s$, $C_u$ as single-layer MLPs. During training, we alternatively train VAEs and $D_T$. We train VAEs first, and after training VAEs $n_d$ times, we train $D_T$ once.

We use the LBFGS implementation from Scikit-learn~\cite{scikit-learn} to train $C_d$ and divide our training set into a validation seen set and a validation unseen set. As the training of $C_d$ requires seen and unseen skeleton features ($f^s_x$, $f^u_x$), we re-train other components using the validation seen set and use the validation unseen set to provide unseen skeleton features to train $C_d$. Finally, the trained $C_d$ is used to make inferences on the testing set. The number of classes in the validation unseen set is the same as the original unseen class set $| \mathcal{Y}_{u} |$.

We use the cyclical annealing schedule~\cite{fu2019cyclical} to train our VAEs because cyclical annealing mitigates the KL divergence vanishing problem. 
At the beginning of each epoch, we set the actual training hyperparameters $\lambda'_2$, $\beta'_1$, and $\beta'_2$ as 0 until we use one-third training samples. Thereafter, we progressively increase $\lambda'_2$, $\beta'_1$, and $\beta'_2$ to $\lambda_2$, $\beta_x$, and $\beta_y$ based on the number of trained samples, e.g,
\begin{equation}
    \lambda'_2=\left\{
    \begin{array}{ll}
    0         & \mbox{if $k<\frac{1}{3}n$}; \\
    \frac{3}{2}(\frac{k}{n}-\frac{1}{3})\lambda_2 & \mbox{if $k\geq \frac{1}{3}n$},\\
    \end{array}
    \right.
\end{equation}
where $k$ and $n$ are the index and total number of training samples in an epoch.
We set $\lambda_1$ as 0 in our first epoch and 1 for all subsequent epochs. We conduct our experiments on a machine equipped with an Intel i7-13700 CPU, an NVIDIA RTX 3090 GPU, and 32GB RAM. We implement our method using PyTorch 2.1.0, scikit-learn 1.3.2, and scipy 1.11.3. It takes 4.6 hours to train our model for a 55/5 split of the NTU RGB+D 60 dataset, and 8.7 hours for a 110/10 split of the NTU RGB+D 120 dataset. We determine the  hyperparameters through random search, as listed in Tables \ref{tab:hyperparameters-sota} and \ref{tab:hyperparameters-optimized}. The hyperparameter search space is detailed in Supplementary Materials Section A.
\begin{table}[t]
    \centering
    \caption{Setting for comparison with existing methods.}
    \begin{tabular}{l c@{\hspace{30pt}} c}
    \toprule
    & NTU-60 & NTU-120\\
    \midrule
    Skeleton Feature Extractor & \multicolumn{2}{c}{Shift-GCN~\cite{cheng2020shiftgcn}}\\
    Text Feature Extractor & \multicolumn{2}{c}{Sentence-BERT~\cite{reimers2019sentence}}\\
    Epochs & \multicolumn{2}{c}{10}\\
    Optimizer & \multicolumn{2}{c}{Adam}\\
    Optimizer Momentum & \multicolumn{2}{c}{$\beta_1=0.9, \beta_2=0.999$} \\
    Batch size & \multicolumn{2}{c}{32}\\
    Learning rate & 3.39e-05 & 3.48e-05\\
    Weights of $D_{\it KL}$ in $\mathcal{L}_{\it VAE}$ & \multicolumn{2}{c}{$\beta_x = 0.023, \beta_y = 0.011$}\\
    Weight of $\mathcal{L}_T$ & \multicolumn{2}{c}{$\lambda_2 = 0.011$} \\
    Discriminator steps $n_d$ & 5 & 4\\
    Hidden dim. of $z^r_x$ and $z_y$ & 160 & 256\\
    Hidden dim. of $z^v_x$ & 8 & 32\\
    \bottomrule
    \end{tabular}
    \label{tab:hyperparameters-sota}
\end{table}
\begin{table}[t]
    \centering
    \caption{ZSL accuracy (\%) on the NTU RGB+D datasets.}
    \begin{tabular}{@{}l*{4}{c}@{}}
        \toprule
        \multirow{2}{*}{Method}              & \multicolumn{2}{c}{NTU-60} & \multicolumn{2}{c}{NTU-120}                                  \\
        \cmidrule(lr){2-3} \cmidrule(lr){4-5}
                                             & {55/5 split}               & {48/12 split}               & {110/10 split} & {96/24 split} \\
        \midrule
        ReViSE~\cite{hubert2017revise}       & 53.91                      & 17.49                       & 55.04          & 32.38         \\
        JPoSE~\cite{wray2019JPoSE}           & 64.82                      & 28.75                       & 51.93          & 32.44         \\
        CADA-VAE~\cite{schonfeld2019cadavae} & 76.84                      & 28.96                       & 59.53          & 35.77         \\
        SynSE~\cite{gupta2021synse}          & 75.81                      & 33.30                       & 62.69          & 38.70         \\
        SMIE~\cite{zhou2023smie}             & 77.98                      & 40.18                       & 65.74          & 45.30         \\
        \midrule
        SA-DVAE                              & {\bf 82.37}                & {\bf 41.38}                 & {\bf 68.77}    & {\bf 46.12}   \\
        \bottomrule
    \end{tabular}
    \label{tab:zsl-sota}
\end{table}
\begin{table}[t]
    \centering
    \caption{GZSL metrics: seen class accuracy \small{$\textit{Acc}_{s}$}, unseen class accuracy \small{$\textit{Acc}_{u}$}, and their harmonic mean \small{$\textit{H}$} (\%) on the NTU RGB+D datasets. *: SynSE paper reports 29.22, but it is a miscalculation.
}
    \resizebox{\linewidth}{!}{
    \begin{tabular}{@{}l*{4}{cca}@{}}
        \toprule
        \multirow{3}{*}{Method}              & \multicolumn{6}{c}{NTU-60}     & \multicolumn{6}{c}{NTU-120}                                                                                                                                                            \\
        \cmidrule(lr){2-7} \cmidrule(lr){8-13}

                                             & \multicolumn{3}{c}{55/5 split} & \multicolumn{3}{c}{48/12 split} & \multicolumn{3}{c}{110/10 split} & \multicolumn{3}{c}{96/24 split}                                                                                   \\
        \cmidrule(lr){2-4} \cmidrule(lr){5-7} \cmidrule(lr){8-10} \cmidrule(lr){11-13}

                                             & \small{$\textit{Acc}_{s}$}     & \small{$\textit{Acc}_{u}$}      & \small{$\textit{H}$}
                                             & \small{$\textit{Acc}_{s}$}     & \small{$\textit{Acc}_{u}$}      & \small{$\textit{H}$}
                                             & \small{$\textit{Acc}_{s}$}     & \small{$\textit{Acc}_{u}$}      & \small{$\textit{H}$}
                                             & \small{$\textit{Acc}_{s}$}     & \small{$\textit{Acc}_{u}$}      & \small{$\textit{H}$}                                                                                                                                 \\
        \midrule
        ReViSE~\cite{hubert2017revise}       & 74.22                          & 34.73                           & 47.32*
               & 62.36                           & 20.77 & 31.16       & 48.69 & 44.84 & 46.68       & 49.66 & 25.06 & 33.31       \\
        JPoSE~\cite{wray2019JPoSE}           & 64.44                          & 50.29                           & 56.49                            & 60.49                           & 20.62 & 30.75       & 47.66 & 46.40 & 47.05       & 38.62 & 22.79 & 28.67       \\
        CADA-VAE~\cite{schonfeld2019cadavae} & 69.38                          & 61.79                           & 65.37                            & 51.32                           & 27.03 & 35.41       & 47.16 & 49.78 & 48.44       & 41.11 & 34.14 & 37.31       \\
        SynSE~\cite{gupta2021synse}          & 61.27                          & 56.93                           & 59.02                            & 52.21                           & 27.85 & 36.33       & 52.51 & 57.60 & 54.94       & 56.39 & 32.25 & 41.04       \\
        \midrule
        SA-DVAE                              & 62.28                          & 70.80                           & {\bf 66.27}                      & 50.20                           & 36.94 & {\bf 42.56} & 61.10 & 59.75 & {\bf 60.42} & 58.82 & 35.79 & {\bf 44.50} \\
        \bottomrule
    \end{tabular}
    }
    \label{tab:gzsl-sota}
\end{table}

\noindent\textbf{Comparison with SOTA methods.}
We compare our method with several state-of-the-art zero-shot action recognition methods using the setting shown in~\Cref{tab:hyperparameters-sota} and report their results in Tables~\ref{tab:zsl-sota} and \ref{tab:gzsl-sota}.
We use the same feature extractors and class splits as the one used by SynSE, and the only difference lies in the network architecture. 

The results show that SA-DVAE works well, in particular for unseen classes.
Furthermore, for the more challenging GZSL task, SA-DVAE even improves more over existing methods. On the NTU RGB+D 60 dataset, SA-DVAE improves the accuracy of (+7.25\% and +6.23\%) in the GZSL protocol, greater than the (+4.39\% and +1.2\%) in the ZSL one.

\begin{table}[t]
    \caption{Settings for the random-split experiment.}
    \centering
    \begin{tabular}{l c@{\hspace{30pt}} c @{\hspace{30pt}}c}
    \toprule
    & NTU-60 & NTU-120 & PKU-MMD\\
    \midrule
    Skeleton Feature Extractor & \multicolumn{3}{c}{ST-GCN~\cite{yan2018stgcn}}\\
    Text Feature Extractor & \multicolumn{3}{c}{CLIP-ViT-B/32~\cite{radford2021clip}}\\
    Epochs & \multicolumn{3}{c}{10}\\
    Optimizer & \multicolumn{3}{c}{Adam}\\
    No. of unseen classes & 5 & 10 & 5 \\
    Optimizer Momentum & \multicolumn{3}{c}{$\beta_1=0.9, \beta_2=0.999$} \\
    Batch size & 32 & 32 & 64\\
    Learning rate & 3.60e-05 & 7.36e-5 & 3.07e-05\\
    Weights of $D_{\it KL}$ in $\mathcal{L}_{\it VAE}$ & \multicolumn{3}{c}{$\beta_x = 0.023, \beta_y = 0.011$}\\
    Weight of $\mathcal{L}_T$ & \multicolumn{3}{c}{$\lambda_2 = 0.011$} \\
    Discriminator steps $n_d$ & 7 & 12 & 2\\
    Hidden dim. of $z^r_x$ and $z_y$ & 224 & 176 & 128\\
    Hidden dim. of $z^v_x$ & 16 & 20 & 16\\
    
    \bottomrule
    \end{tabular}
    \label{tab:hyperparameters-optimized}
\end{table}

\noindent\textbf{Random Class Splits and Improved Feature Extractors.}
The setting of class splits is crucial for accuracy calculation and Tables~\ref{tab:zsl-sota} and \ref{tab:gzsl-sota} only show results of a few predefined splits, which can not infer the overall performance on a complete dataset. Thus, we follow Zhou \etal's approach~\cite{zhou2023smie} to randomly select several unseen classes as a new split, repeat it three times, and report the average performance.
In addition, we use improved skeleton feature extractor ST-GCN~\cite{yan2018stgcn} and text extractor CLIP~\cite{radford2021clip}, chosen for their broad applicability and robust performance across different domains.
We also tested different feature extractors, which can be found in Supplementary Materials Section B.

\begin{table}[t]
    \centering
    \caption{Average ZSL accuracy (\%) under the random split setting on the NTU-60, NTU-120, and PKU-MMD datasets. FD: feature disentanglement. TC: adversarial total correlation penalty. \dag: PoS tags for the PKU-MMD dataset are obtained from spaCy~\cite{honnibal2020spacy}.}
    \begin{tabular}{@{}l*{3}{c}@{}}
        \toprule
        \multirow{2}{*}{Method}              & {NTU-60}     & {NTU-120}      & {PKU-MMD}    \\
                                             & {55/5 split} & {110/10 split} & {46/5 split} \\
        \midrule
        ReViSE~\cite{hubert2017revise}       & 60.94        & 44.90          & 59.34        \\
        JPoSE\dag~\cite{wray2019JPoSE}       & 59.44        & 46.69          & 57.17        \\
        CADA-VAE~\cite{schonfeld2019cadavae} & 61.84        & 45.15          & 60.74        \\
        SynSE\dag~\cite{gupta2021synse}      & 64.19        & 47.28          & 53.85        \\
        SMIE~\cite{zhou2023smie}             & 65.08        & 46.40          & 60.83        \\
        \midrule
        Naive alignment                      & 69.26        & 39.73          & 60.13        \\
        FD                                   & 82.21        & 49.18          & 60.97        \\
        SA-DVAE (FD+TC)                      & {\bf 84.20}  & {\bf 50.67}    & {\bf 66.54}  \\
        \bottomrule
    \end{tabular}
    \label{tab:zsl-optimized}
\end{table}
\begin{table}[!ht]
    \centering
    \caption{Average GZSL metrics: seen class accuracy \small{$\textit{Acc}_{s}$}, unseen class accuracy \small{$\textit{Acc}_{u}$}, and their harmonic mean \small{$\textit{H}$} (\%) under the random split setting on the NTU-60, NTU-120, and PKU-MMD datasets. FD: feature disentanglement. TC: adversarial total correlation penalty. \dag: PoS tags for the PKU-MMD dataset are obtained from spaCy~\cite{honnibal2020spacy}.}
    \begin{tabular}{@{}l*{3}{cca}@{}}
        \toprule
        \multirow{3}{*}{Method}              & \multicolumn{3}{c}{NTU-60}      & \multicolumn{3}{c}{NTU-120}      & \multicolumn{3}{c}{PKU-MMD}                                                                \\

                                             & \multicolumn{3}{c}{55/5 splits} & \multicolumn{3}{c}{110/10 split} & \multicolumn{3}{c}{46/5 split}                                                             \\
        \cmidrule(lr){2-4} \cmidrule(lr){5-7} \cmidrule(lr){8-10}

                                             & \small{$\textit{Acc}_{s}$}      & \small{$\textit{Acc}_{u}$}       & \small{$\textit{H}$}
                                             & \small{$\textit{Acc}_{s}$}      & \small{$\textit{Acc}_{u}$}       & \small{$\textit{H}$}
                                             & \small{$\textit{Acc}_{s}$}      & \small{$\textit{Acc}_{u}$}       & \small{$\textit{H}$}                                                                       \\
        \midrule
        ReViSE~\cite{hubert2017revise}       & 71.75                           & 52.06                            & 60.34                          & 48.29 & 34.64 & 40.34       & 60.89 & 42.16 & 49.82       \\
        JPoSE \dag~\cite{wray2019JPoSE}      & 66.25                           & 54.92                            & 60.05                          & 49.43 & 39.14 & 43.69       & 60.26 & 45.18 & 51.64       \\
        CADA-VAE~\cite{schonfeld2019cadavae} & 77.35                           & 58.14                            & 66.38                          & 51.09 & 41.24 & 45.64       & 63.17 & 35.86 & 45.75       \\
        SynSE \dag~\cite{gupta2021synse}     & 75.84                           & 60.77                            & 67.47                          & 41.73 & 45.36 & 43.47       & 63.09 & 40.69 & 49.47       \\
        \midrule
        Naive alignment                  & 82.11                           & 47.99                            & 60.58                          & 57.01 & 31.62 & 40.68       & 58.76 & 43.14 & 49.75       \\
        FD                               & 82.31                           & 61.98                            & 70.71                          & 58.57 & 37.83 & 45.97       & 58.11 & 48.15 & 52.66       \\
        SA-DVAE (FD+TC)                  & 78.16                           & 72.60                            & {\bf 75.27}                    & 58.09 & 40.23 & {\bf 47.54} & 58.49 & 51.40 & {\bf 54.72} \\
        \bottomrule
    \end{tabular}
    \label{tab:gzsl-optimized}
\end{table}

\begin{figure}[t]
    \centering
    \noindent\resizebox{\textwidth}{!}{
    \includegraphics[]{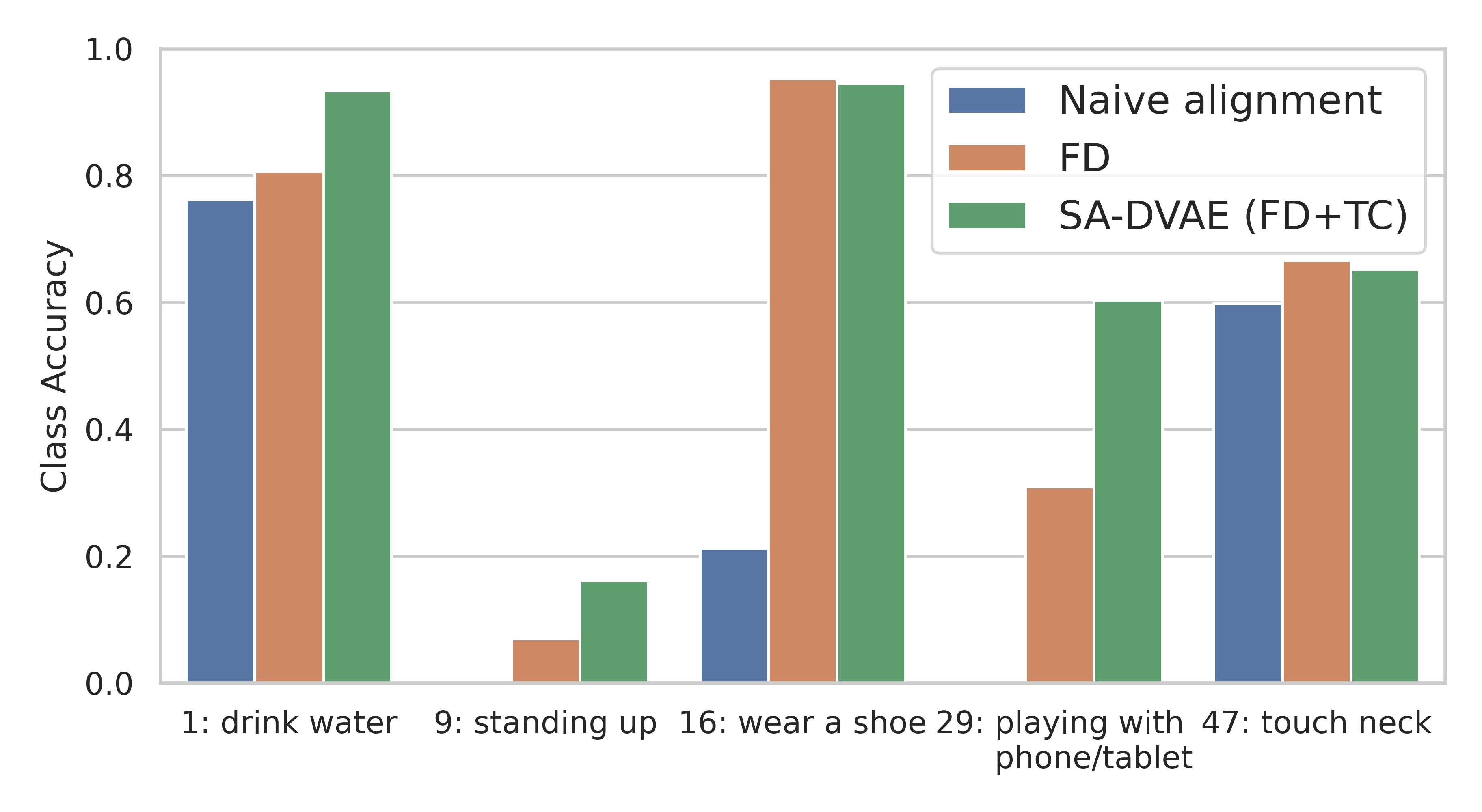}}
    \caption{Unseen per-class accuracy of the NTU-60 dataset. The unseen split \{1, 9, 16, 29, 47\} is used in a challenging run of our random-split GZSL experiments.}
    \label{fig:unseen-per-class-acc-optimized}
\end{figure}

\begin{table}[]
    \centering
    \caption{Average GZSL metrics (\%) of different seen classifier input under the random split setting on the NTU-60, NTU-120, and PKU-MMD datasets.}
    \begin{tabular}{@{}l*{3}{cca}@{}}
        \toprule
        \multirow{3}{*}{Method}              & \multicolumn{3}{c}{NTU-60}      & \multicolumn{3}{c}{NTU-120}      & \multicolumn{3}{c}{PKU-MMD}                                                                \\

                                             & \multicolumn{3}{c}{55/5 splits} & \multicolumn{3}{c}{110/10 split} & \multicolumn{3}{c}{46/5 split}                                                             \\
        \cmidrule(lr){2-4} \cmidrule(lr){5-7} \cmidrule(lr){8-10}

                                             & \small{$\textit{Acc}_{s}$}      & \small{$\textit{Acc}_{u}$}       & \small{$\textit{H}$}
                                             & \small{$\textit{Acc}_{s}$}      & \small{$\textit{Acc}_{u}$}       & \small{$\textit{H}$}
                                             & \small{$\textit{Acc}_{s}$}      & \small{$\textit{Acc}_{u}$}       & \small{$\textit{H}$}                                                                       \\
        \midrule
        SA-DVAE ($z^r_x$ as input)  & 72.00   & 71.48   & 71.74 & 55.35 & 39.00 & 45.76 & 59.16 & 49.73 & 54.04 \\
        SA-DVAE ($f_x$ as input)    & 78.16 & 72.60 & {\bf 75.27}   & 58.09 & 40.23 & {\bf 47.54} & 58.49 & 51.40 & {\bf 54.72} \\
        \bottomrule
    \end{tabular}
    \label{tab:gzsl-fx-zx-comparison}
\end{table}

\Cref{tab:hyperparameters-optimized} shows our settings and Tables~\ref{tab:zsl-optimized} and \ref{tab:gzsl-optimized} show the results, where naive alignment means that we disable $D_T$ and remove the extra head for $z^v_x$, and FD means that we disable $D_T$. The results show that both feature disentanglement and total correlation penalty contribute to accuracy improvements, and feature disentanglement is the major contributor, \eg, +12.95\% on NTU-60 compared to naive alignment in Table~\ref{tab:zsl-optimized}. 
The adversarial total correlation penalty (TC) slightly reduces the accuracy for seen classes but significantly improves unseen and overall accuracy. This is because TC enhances the embedding quality by reducing feature redundancy, making the domain classifier less biased towards seen classes. Consequently leading to improved generalization. The results in \cref{tab:gzsl-optimized} highlight this trade-off, where the improved harmonic mean indicates a more balanced and robust performance across both seen and unseen classes.

\begin{figure}[t]
    \centering
    \noindent\resizebox{\textwidth}{!}{
    \includegraphics[]{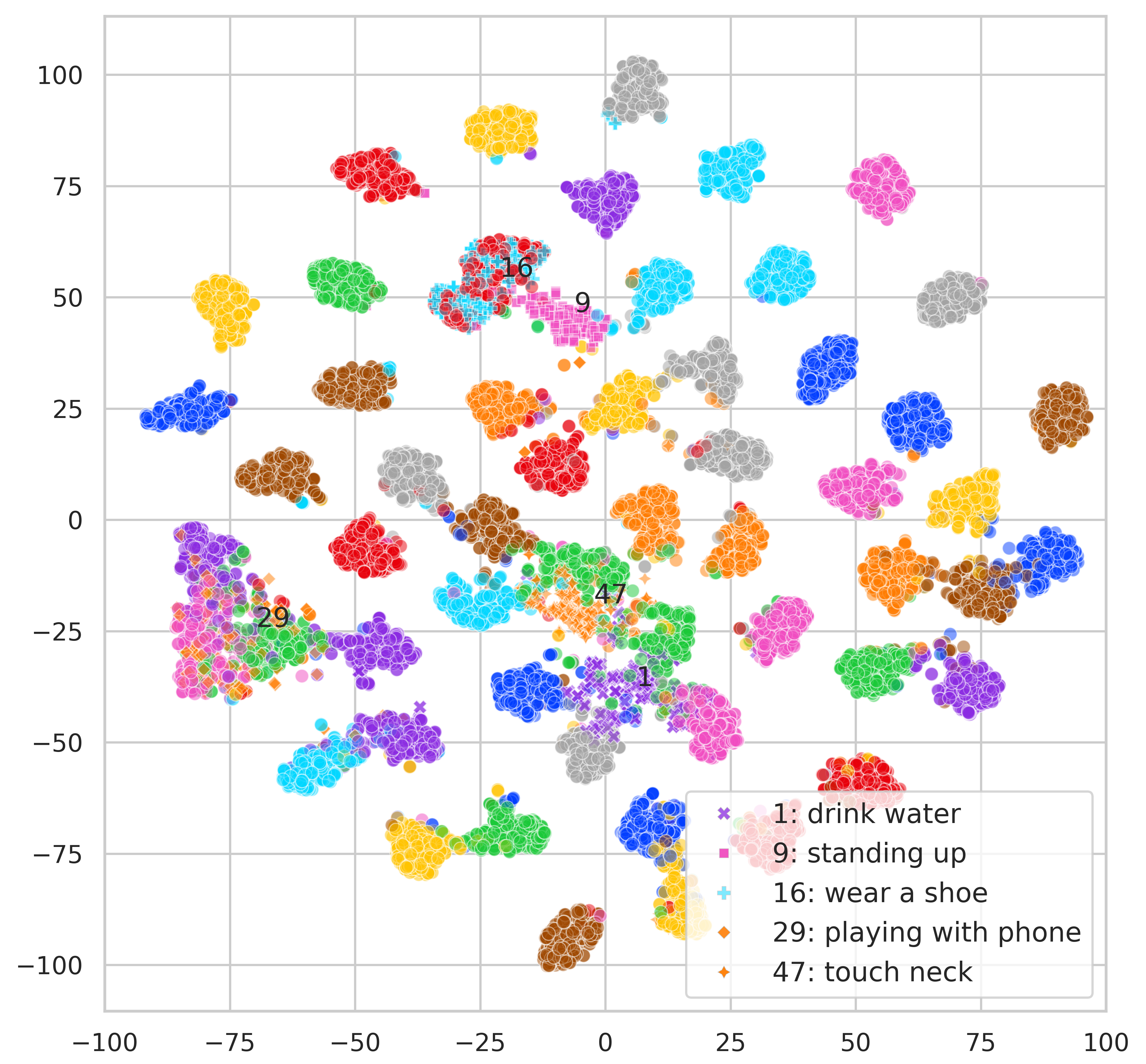}}
    \caption{t-SNE visualization of $f_x$ of the NTU-60 dataset. The unseen split \{1, 9, 16, 29, 47\} is used in a run of our random-split GZSL experiments. Best viewed in color.}
    \label{fig:tsne-vis-optimized}
\end{figure}

From our three runs of the random-split experiment on the NTU-60 dataset (average results is shown in Table~\ref{tab:zsl-optimized}), we pick the most challenging run and show its per-class accuracy in \cref{fig:unseen-per-class-acc-optimized} and the t-SNE visualization of skeleton features ($f_x$) in \cref{fig:tsne-vis-optimized}.
The labels of classes 16 and 17 are ``wear a shoe'' and ``take off a shoe'' and their movements are acted as a person sitting on a chair who bends down her upper body and stretches her arm to touch her shoe. The skeleton sequences of the two classes are highly similar so are their extracted features. In \cref{fig:tsne-vis-optimized}, samples of classes 16 and 17 are overlapped, and naive alignment generates poor accuracy on class 16. Similarly, naive alignment generates near-zero accuracy on classes 9 and 29. Since both classes 9 and 16 share similar skeleton sequences and were unseen during training, their features appear highly similar. This similarity leads naive alignment to misclassify samples belonging to class 9 as class 16. We can see significant improvements with the addition of FD and TC. 
These techniques allow the model to prioritize semantic-related information and improve classification performance.

\noindent \textbf{Impact of Replacing Skeleton Feature $f_x$ with Semantic-Related Latent Vector $z^r_x$ in Seen Classifier}
We replace the input skeleton feature $f_x$ of the seen classifier with the disentangled semantic-related latent vector $z^r_x$ under the random-split setting listed in \cref{tab:hyperparameters-optimized} and report results in \cref{tab:gzsl-fx-zx-comparison}.
Notably, since the semantic-irrelevant terms also contain information that is beneficial for classification but not necessary related to the text descriptions, $f_x$ retains both semantic-related and irrelevant details. This dual retention enhances performance compared to $z^r_x$, which focuses solely on semantic-related information.

We incorporate zero-shot learning and action recognition techniques, including pose canonicalization~\cite{holden2016deep} and enhanced action descriptions~\cite{zhou2023smie}, with additional experimental results in Supplementary Materials Section C.

\section{Conclusion}
\label{sec:conclusion}
ZSL study aims to leverage knowledge from one domain to help solve problems in another domain and has been proven useful for action recognition tasks, in particular for 3D skeleton data because it is expensive and labor-consuming to build accurately labeled datasets.
Although there are several existing methods in the literature, they never address the asymmetry problem between skeleton data and text description.
In this paper, we propose SA-DVAE, a cross-modality alignment model using the feature disentanglement approach to differentiate skeleton data into two independent representations, the semantic-related and irrelevant ones.
Along with an adversarial discriminator to enhance the feature disentanglement, our experiments show that the proposed method generates better performance over existing methods on three benchmark datasets in both ZSL and GZSL protocols.

\section*{Acknowledgments}
This research was supported by the National Science and Technology Council of Taiwan under grant number 111-2622-8-002-028. The authors would like to thank the NSTC for its generous support.

\bibliographystyle{splncs04}
\bibliography{ref}

\clearpage
\setcounter{section}{0}
\setcounter{table}{0}
\setcounter{figure}{0}
\renewcommand{\thesection}{\Alph{section}}
\renewcommand{\thesubsection}{\thesection.\arabic{subsection}}
\renewcommand{\thetable}{\Alph{table}}

\section{Hyperparameter Search Space and Sensitivity}
We show our search space and initial values in \cref{tab:search_space1}.
\begin{table}[h]
    \centering
    \caption{Hyperparameter search space and initial values}
    \begin{tabular}{l@{\hspace{2pt}}lcc}
        \toprule
         No. & Hyperparameter & Space & Initial \\
         \midrule
         1 & $\beta_x$ and $\beta_y$ & (0, 1.0) & \\
         2 & Learning rate's exponent & (-6.0, -3.0) & -5 \\
         3 & Batch size & \{32, 64, 128, 256\} &  64 \\
         4 & Discriminator steps $n_d$ & \{1, 2, 3, ..., 16\} & 10\\
         5 & Hidden dim. of $z^r_x$ and $z_y$ & \{128, 144, 160, ..., 256\} & 192\\
         6 & Hidden dim. of $z^v_x$ & \{8, 12, 16, ..., 32\} & 8 \\
         \bottomrule
    \end{tabular}
    \label{tab:search_space1}
\end{table}

We first fix No. 2$\sim$6 and randomly sample No. 1 in uniform distribution 5 times. We choose the one generating the highest GZSL harmonic mean on the validation set. Then we fix No. 1 and randomly sample No. 2$\sim$6 100 times.

\Cref{tab:betaxy_ntu60_1} shows the influence of $\beta_x$ and $\beta_y$ on the experiments of Tables 6 and 7 in the main paper. As reported in Table 5 in the main paper, we use $\beta_x$ as 0.023 and $\beta_y$ as 0.011 because they perform best on the validation set. We leave out $\beta_x$ and $\beta_y$ $\ge$ 0.2 because their performance is low.

\begin{table}[h]
    \centering
    \caption{Sensitivity of $\beta_x$ and $\beta_y$ on ZSL and GZSL metrics.}
    \begin{tabular}{ccacca}
        \toprule
        \multirow{2}{*}{$\beta_x$} & \multirow{2}{*}{$\beta_y$} & \multicolumn{1}{c}{ZSL} & \multicolumn{3}{c}{GZSL} \\
        \cmidrule(lr){3-3} \cmidrule(lr){4-6}
        & & \small{$\textit{Acc}$} &\small{$\textit{Acc}_{s}$}      & \small{$\textit{Acc}_{u}$}       & \small{$\textit{H}$} \\
        \midrule
        0.010 & 0.010 & 83.60 & 81.14 & 67.14 & 73.48 \\
        0.050 & 0.010 & \textbf{84.39} & 73.87 & 73.24 & 73.55 \\
        0.010 & 0.050 & 83.13 & 77.61 & 71.22 & 74.28 \\
        0.050 & 0.050 & 83.62 & 70.74 & 74.23 & 72.44 \\
        0.100 & 0.100 & 82.79 & 75.96 & 68.86 & 72.24 \\
        0.200 & 0.200 & 76.12 & 71.79 & 62.90 & 67.05 \\
        \midrule
        0.023 & 0.011 & 84.20 & 78.16 & 72.60 & \textbf{75.27} \\
        \bottomrule
    \end{tabular}
    \label{tab:betaxy_ntu60_1}
    \vspace{-4pt}
\end{table}

\section{Feature Extractors}
We show an example by re-organizing Tables 6 and 7 in the main paper as \cref{tab:feature_extractors_1}. Their dataset, splits, and hyperparameters are the same and the only difference lies in feature extractors. 
Experimental results show that extractors matter and our proposed ST-GCN+CLIP works best.

\begin{table}[h]
    \centering
    \caption{Average ZSL accuracy and GZSL metrics (\%) of different feature extractors under the random split setting on NTU-60.}
    \begin{tabular}{lacca}
        \toprule
        \multirow{2}{*}{Feature Extractors}  & \multicolumn{1}{c}{ZSL} & \multicolumn{3}{c}{GZSL} \\
        \cmidrule(lr){2-2} \cmidrule(lr){3-5}
        & \small{$\textit{Acc}$} &\small{$\textit{Acc}_{s}$}      & \small{$\textit{Acc}_{u}$}       & \small{$\textit{H}$} \\
        \midrule
        ST-GCN~\cite{yan2018stgcn} + Sentence-BERT~\cite{reimers2019sentence} & 74.38 & 71.39 & 61.02 & 65.80 \\
        PoseC3D~\cite{duan2022revisiting} + CLIP~\cite{radford2021clip} & 81.84 & 83.48 & 66.89 & 74.27 \\
        \midrule
        ST-GCN~\cite{yan2018stgcn} + CLIP~\cite{radford2021clip} & \textbf{84.20} & 78.16 & 72.60 & \textbf{75.27} \\
        \bottomrule
    \end{tabular}
    \label{tab:feature_extractors_1}
\end{table}

\section{Combining with Existing Methods}
To potentialy improve our performance, we combine our method with pose canonicalization on skeleton data~\cite{holden2016deep} and enhanced class descriptions by a large language model proposed in SMIE~\cite{zhou2023smie}. We will discuss the details and experimental results in the following sections.

\subsection{Pose Canonicalization on Skeleton Data}
The difference in the forward direction of the skeleton data introduces additional noise into the training process. Therefore, we implement the method proposed by Holden \etal~\cite{holden2016deep} to canonicalize the skeleton data by rotating them so that they face the same direction. We compute the cross product between the vertical axis and the average vector of the left and right shoulders and hips to determine the new forward direction of the body. We then apply a rotation matrix to canonicalize the pose.

Tables \ref{tab:zsl-optimized-canonicalized} and \ref{tab:gzsl-optimized-canonicalized} present the experimental results under random split settings listed in Table 5 of the main paper. In zero-shot settings, we observe that canonicalization of skeleton data has little effect on model performance. For generalized zero-shot settings, we note a slight decrease in both seen and unseen accuracies. We hypothesize that this is because canonicalization reduces the variation in the skeleton dataset. This reduction in diversity limits the range of examples the model encounters during training, which may ultimately impair its ability to generalize effectively.

\begin{table}[h]
    \centering
    \caption{Average ZSL accuracy (\%) under the random split setting on the NTU-60, NTU-120, and PKU-MMD datasets.}
    \begin{tabular}{@{}l*{3}{c}@{}}
        \toprule
        \multirow{2}{*}{Method}              & {NTU-60}     & {NTU-120}      & {PKU-MMD}    \\
                                             & {55/5 split} & {110/10 split} & {46/5 split} \\
        \midrule
        SA-DVAE                              & {\bf 84.20}  & {\bf 50.67}    & 66.54        \\
        SA-DVAE + pose canonicalization              & 84.03        & 50.04          & {\bf 67.56}  \\ 
        \bottomrule
    \end{tabular}
    \label{tab:zsl-optimized-canonicalized}
\end{table}

\begin{table}[h]
    \centering
    \caption{Average GZSL metrics: seen class accuracy \small{$\textit{Acc}_{s}$}, unseen class accuracy \small{$\textit{Acc}_{u}$}, and their harmonic mean \small{$\textit{H}$} (\%) under the random split setting on the NTU-60, NTU-120, and PKU-MMD datasets.}
    \begin{tabular}{@{}l*{3}{cca}@{}}
        \toprule
        \multirow{3}{*}{Method}              & \multicolumn{3}{c}{NTU-60}      & \multicolumn{3}{c}{NTU-120}      & \multicolumn{3}{c}{PKU-MMD}                                                                \\

                                             & \multicolumn{3}{c}{55/5 splits} & \multicolumn{3}{c}{110/10 split} & \multicolumn{3}{c}{46/5 split}                                                             \\
        \cmidrule(lr){2-4} \cmidrule(lr){5-7} \cmidrule(lr){8-10}

                                             & \small{$\textit{Acc}_{s}$}      & \small{$\textit{Acc}_{u}$}       & \small{$\textit{H}$}
                                             & \small{$\textit{Acc}_{s}$}      & \small{$\textit{Acc}_{u}$}       & \small{$\textit{H}$}
                                             & \small{$\textit{Acc}_{s}$}      & \small{$\textit{Acc}_{u}$}       & \small{$\textit{H}$}                                                                       \\
        \midrule
        SA-DVAE                          & 78.16                           & 72.60                            & {\bf 75.27}                    & 58.09 & 40.23 & {\bf 47.54} & 58.49 & 51.40 & {\bf 54.72} \\
        SA-DVAE + pose canonicalization          & 72.84                           & 69.85                            & 71.31                          & 56.78 & 35.22 & 43.47       & 54.13 & 50.60 & 52.30 \\
        \bottomrule
    \end{tabular}
    \label{tab:gzsl-optimized-canonicalized}
\end{table}

\subsection{Enhanced Class Descriptions by a Large Language Model (LLM)}
Zhou \etal~\cite{zhou2023smie} propose to use an LLM to augment class descriptions with richer action-related information and we directly compare our and their methods by using their augmented descriptions.
We report results using the same setting for random split and list our hyperparameters in Table~\ref{tab:hyperparameters-llm}, and generate results shown in Tables~\ref{tab:zsl-chat} and \ref{tab:gzsl-chat}, which show that SA-DAVE outperforms SMIE using augmented descriptions in both ZSL and GZSL protocols and LLM-augmented descriptions significantly improve unseen accuracy while marginally decreasing seen accuracy. This is consistent with the pattern observed in the ablation study, indicating that the models achieve a more balanced prediction with minimal bias toward seen or unseen classes.

\begin{table}[t]
    \caption{Settings for LLM-augmented class descriptions under the random split setting.}
    \centering
    \begin{tabular}{l c@{\hspace{30pt}} c}
    \toprule
    & NTU-60 & NTU-120\\
    \midrule
    Skeleton Feature Extractor & \multicolumn{2}{c}{ST-GCN~\cite{yan2018stgcn}}\\
    Text Feature Extractor & \multicolumn{2}{c}{CLIP-ViT-B/32~\cite{radford2021clip}}\\
    Epochs & \multicolumn{2}{c}{10}\\
    Optimizer & \multicolumn{2}{c}{Adam}\\
    No. of unseen classes & 5 & 10 \\
    Optimizer Momentum & \multicolumn{2}{c}{$\beta_1=0.9, \beta_2=0.999$} \\
    Batch size & 32 & 24\\
    Learning rate & 4.94e-05 & 2.13e-05\\
    Weights of $D_{\it KL}$ in $\mathcal{L}_{\it VAE}$ & \multicolumn{2}{c}{$\beta_x = 0.023, \beta_y = 0.011$}\\
    Weight of $\mathcal{L}_T$ & \multicolumn{2}{c}{$\lambda_2 = 0.011$} \\
    Discriminator steps $n_d$ & 4 & 16\\
    Hidden dim. of $z^r_x$ and $z_y$ & 96 & 304\\
    Hidden dim. of $z^v_x$ & 8 & 12\\
    
    \bottomrule
    \end{tabular}
    \label{tab:hyperparameters-llm}
\end{table}

\begin{table}[]
    \centering
    \caption{ZSL accuracy (\%) with LLM-augmented class descriptions on the NTU-60 and NTU-120 datasets.}
    \label{tab:zsl-chat}
    \begin{tabular}{lcc}
        \toprule
        \multirow{2}{*}{Method}              & {NTU-60}     & {NTU-120} \\
                                             & {55/5 split} & {110/10 split} \\
        \midrule
        SMIE~\cite{zhou2023smie} & 65.08 & 46.40 \\
        SMIE + augmented text~\cite{zhou2023smie} & 70.89 & 52.04 \\
        \midrule
        SA-DVAE & 84.20 & 50.67 \\
        SA-DVAE + augmented text & {\bf 87.61} & {\bf 57.16}\\
        \bottomrule
    \end{tabular}
\end{table}

\begin{table}[]
    \centering
    \caption{GZSL metrics (\%) with LLM-augmented class descriptions on the NTU-60 and NTU-120 datasets.}
    \label{tab:gzsl-chat}
    \begin{tabular}{lccacca}
        \toprule
        \multirow{3}{*}{Method}              & \multicolumn{3}{c}{NTU-60}      & \multicolumn{3}{c}{NTU-120} \\

                                             & \multicolumn{3}{c}{55/5 splits} & \multicolumn{3}{c}{110/10 split} \\
        \cmidrule(lr){2-4} \cmidrule(lr){5-7}

                                             & \small{$\textit{Acc}_{s}$}      & \small{$\textit{Acc}_{u}$}       & \small{$\textit{H}$}
                                             & \small{$\textit{Acc}_{s}$}      & \small{$\textit{Acc}_{u}$}       & \small{$\textit{H}$} \\
        \midrule
        SA-DVAE & 78.16 & 72.60 & 75.27 & 58.09 & 40.23 & 47.54 \\
        SA-DVAE + augmented text & 74.54 & 76.50 & {\bf 75.51} & 53.32 & 48.36 & {\bf 50.72}\\
        \bottomrule
    \end{tabular}
\end{table}

\end{document}